# EnergyPatchTST: Multi-scale Time Series Transformers with Uncertainty Estimation for Energy Forecasting*

Wei Li[1] [0009-0008-8108-4854], Zixin Wang[1][0009-0006-8815-9279], Qizheng Sun[2][0009-0009-9361-8005], Qixiang Gao[3][0009-0002-1038-6886], and Fenglei Yang[1(✉)]

[1] School of Computer Engineering and Science, Shanghai University, Shanghai, China
liwei008009@163.com, 2494546924@qq.com, flyang@shu.edu.cn
[2] College of Sciences, Shanghai University, Shanghai, China
qizhengsun@126.com
[3] School of Mathematical Sciences, Fudan University, Shanghai, China
3380274918@qq.com

**Abstract.** Accurate and reliable energy time series prediction is of great significance for power generation planning and allocation. At present, deep learning time series prediction has become the mainstream method. However, the multi-scale time dynamics and the irregularity of real data lead to the limitations of the existing methods. Therefore, we propose EnergyPatchTST, which is an extension of the Patch Time Series Transformer specially designed for energy forecasting. The main innovations of our method are as follows: (1) multi-scale feature extraction mechanism to capture patterns with different time resolutions; (2) probability prediction framework to estimate uncertainty through Monte Carlo elimination; (3) integration path of future known variables (such as temperature and wind conditions); And (4) Pre-training and Fine-tuning examples to enhance the performance of limited energy data sets. A series of experiments on common energy data sets show that EnergyPatchTST is superior to other commonly used methods, the prediction error is reduced by 7-12%, and reliable uncertainty estimation is provided, which provides an important reference for time series prediction in the energy field.

## 1 Introduction

Energy time series prediction plays a vital role in modern society, especially in power distribution and renewable energy power generation industry [1]. However, energy forecasting is limited by the complex mode of energy time series that spans multiple time scales, and is often highly dependent on weather conditions and other external factors, especially the production of renewable energy. As the industrial artery, the energy system requires not only point prediction, but also more reliable and confident uncertainty estimation.

Recent advances in deep learning have led to significant improvements in time series forecasting [2]. In particular, Transformer-based architectures [3] have demonstrated impressive performance by effectively capturing long-range dependencies [4,5]. Among these, the Patch Time Series Transformer (PatchTST) [6] has emerged as a state-of-the-art approach by treating subseries as "patches" and leveraging self-attention mechanisms to model dependencies between these patches.

While PatchTST provides a strong foundation for time series modeling, it was not specifically designed for the unique challenges of energy forecasting. Energy time series require models that can effectively:

- Capture patterns at multiple temporal scales, from immediate fluctuations to daily and seasonal patterns
- Incorporate known future information such as weather forecasts
- Provide reliable uncertainty estimates for risk-aware decision-making
- Perform well even with limited training data, which is common in newer energy installations

To address these challenges, we present EnergyPatchTST, an extension of PatchTST specifically designed for energy time series forecasting. Our model introduces four key innovations:

1. **Multi-scale Feature Extraction**: A hierarchical architecture that processes time series at different temporal resolutions, capturing both short-term fluctuations and long-term trends
2. **Uncertainty Estimation**: A Monte Carlo dropout mechanism that provides probabilistic forecasts with calibrated prediction intervals
3. **Future Variables Integration**: A specialized pathway for incorporating known future variables, such as temperature and wind speed forecasts
4. **Pre-training and Fine-tuning**: A transfer learning approach that leverages general time series datasets for pre-training before fine-tuning on specific energy datasets

Experiments on energy datasets demonstrate that EnergyPatchTST outperforms all baselines, reducing forecasting error by 7-12% while providing reliable uncertainty estimates. Our ablation studies confirm that our proposed components contributes significantly to the performance, with the multi-scale feature extraction and uncertainty estimation providing the largest improvements for long-horizon predictions.

The main contributions of this paper are:

- A novel multi-scale architecture for energy time series forecasting that effectively captures patterns at different temporal resolutions
- A probabilistic forecasting framework with uncertainty estimation that provides reliable prediction intervals
- An effective mechanism for integrating future known variables to improve forecast accuracy

- A pre-training and fine-tuning paradigm that enables effective transfer learning for energy forecasting
- Comprehensive experiments demonstrating the effectiveness of our approach on multiple energy forecasting benchmarks

## 2     Related Work

### 2.1    Deep Learning for Time Series Forecasting

Time series forecasting has been revolutionized by deep learning approaches in recent years. Recurrent Neural Networks (RNNs) and their variants such as Long Short-Term Memory (LSTM) [7] and Gated Recurrent Units (GRU) [8] were early adaptations that showed promise for capturing temporal dependencies. DeepAR [9] used autoregressive recurrent networks for probabilistic forecasting. Temporal Convolutional Networks (TCNs) [10] demonstrated competitive performance with more efficient parallel computation.

The introduction of Transformer architectures [3] led to further advances in time series forecasting. Informer [4] addressed the quadratic complexity of self-attention with a ProbSparse mechanism. Autoformer [5] introduced an auto-correlation mechanism for discovering periodicities. FEDformer [11] incorporated frequency domain information through Fourier transformations. Most recently, PatchTST [6] adapted the vision transformer approach by segmenting time series into patches and achieved state-of-the-art performance on many benchmarks.

### 2.2    Energy Forecasting

Energy forecasting presents unique challenges due to its multi-scale nature and dependence on external factors. Traditional approaches included statistical methods such as ARIMA and exponential smoothing [1], while more recent work has focused on machine learning and deep learning techniques.

For renewable energy forecasting, specialized models have been developed to incorporate weather variables and capture complex patterns. Models such as DeepRenewables [12] have demonstrated the value of incorporating domain knowledge into deep learning architectures. Hong et al. [13] provided a comprehensive review of probabilistic energy forecasting, highlighting the importance of uncertainty quantification in this domain.

### 2.3    Uncertainty Estimation in Deep Learning

Uncertainty estimation is particularly important for energy forecasting, where reliable prediction intervals can inform risk-aware decision-making. In deep learning, uncertainty is typically categorized as aleatoric (data uncertainty) or epistemic (model uncertainty) [14].

Monte Carlo dropout [15] provides a simple yet effective approach for estimating epistemic uncertainty by performing multiple forward passes with dropout active during inference. Deep Ensemble methods [16] train multiple models to capture model uncertainty. Recent work has integrated these approaches into time series forecasting, with methods such as DeepAR [9] and N-BEATS [17] providing probabilistic forecasts.

### 2.4   Transfer Learning for Time Series

Transfer learning has shown great success in computer vision and natural language processing, but its application to time series remains less explored. Recent work has demonstrated the potential of pre-training on large, diverse time series datasets before fine-tuning on specific tasks [18]. Approaches such as TL-ESN [19] and Meta-Learning for Time Series [20] have shown promise in this area. Our approach extends these ideas with a specific focus on the unique characteristics of energy time series.

## 3   Methodology

### 3.1   Problem Formulation

Given a historical multivariate time series $\mathbf{X} = \{\mathbf{x}_1, \mathbf{x}_2, \ldots, \mathbf{x}_T\} \in \mathbb{R}^{T \times D}$, where $T$ is the sequence length and $D$ is the feature dimension, the goal of forecasting is to predict future values $\mathbf{Y} = \{\mathbf{x}_{T+1}, \mathbf{x}_{T+2}, \ldots, \mathbf{x}_{T+H}\} \in \mathbb{R}^{H \times D}$ for a horizon $H$. For energy forecasting, we also have access to future known variables $\mathbf{Z} = \{\mathbf{z}_{T+1}, \mathbf{z}_{T+2}, \ldots, \mathbf{z}_{T+H}\} \in \mathbb{R}^{H \times E}$, such as weather forecasts, where $E$ is the dimension of external features.

Instead of providing only point forecasts, we aim to generate probabilistic forecasts with prediction intervals. We denote our model as $f_\theta$, where $\theta$ represents the model parameters. The model outputs both the predicted mean $\widehat{\mathbf{Y}}$ and variance $\widehat{\mathbf{\Sigma}}$:

$$\widehat{\mathbf{Y}}, \widehat{\mathbf{\Sigma}} = f_\theta(\mathbf{X}, \mathbf{Z}) \tag{1}$$

### 3.2   Model Architecture

Our EnergyPatchTST model builds upon the PatchTST architecture [6] with several key enhancements for energy forecasting. Figure 1 provides an overview of our model architecture.

**Multi-scale Feature Extraction** Energy time series exhibit patterns at multiple temporal scales: immediate fluctuations due to local conditions, daily patterns related to human activity or solar cycles, weekly patterns of consumption, and seasonal trends. To capture these multi-scale dynamics, we process the input time series at different temporal resolutions.

Given an input time series $\mathbf{X} \in \mathbb{R}^{T \times D}$, we create multiple scale representations:

$$\mathbf{X}^{(s)} = \text{ScaleTransform}_s(\mathbf{X}), \quad s \in \{1, 2, \ldots, S\} \tag{2}$$

where $s$ represents the scale level, and $S$ is the total number of scales. For scale $s = 1$, we keep the original time series unchanged: $\mathbf{X}^{(1)} = \mathbf{X}$. For scales $s > 1$, we apply averaging over increasingly larger windows:

$$\mathbf{X}^{(s)}_{i,j} = \frac{1}{w_s} \sum_{k=(i-1) \cdot w_s + 1}^{i \cdot w_s} \mathbf{X}_{k,j} \tag{3}$$

where $w_s$ is the window size for scale $s$. In our implementation, we typically use three scales with window sizes $w_1 = 1$ (original), $w_2 = 24$ (daily), and $w_3 = 168$ (weekly) for hourly data. Each scale representation $\mathbf{X}^{(s)}$ is then processed by a separate branch of the model. This parallel processing allows the model to capture patterns at different temporal resolutions simultaneously.

**Patch-based Transformer Encoder** Following the PatchTST approach [6], we segment each scale representation into patches. For a time series $\mathbf{X}^{(s)}$ of length $L_s$, we create patches of length $P$ with stride $\tau$:

$$\mathbf{P}^{(s)} = \text{Patchify}(\mathbf{X}^{(s)}, P, \tau) \tag{4}$$

where $\mathbf{P}^{(s)} \in \mathbb{R}^{N_s \times P \times D}$, and $N_s = \lfloor (L_s - P)/\tau + 1 \rfloor$ is the number of patches.

These patches are then flattened and linearly projected to obtain patch embeddings:

$$\mathbf{E}^{(s)} = \text{Projection}\left(\text{Flatten}(\mathbf{P}^{(s)})\right) \tag{5}$$

where $\mathbf{E}^{(s)} \in \mathbb{R}^{N_s \times d_{model}}$ and $d_{model}$ is the embedding dimension.

The patch embeddings are then processed by a transformer encoder with self-attention mechanisms:

$$\mathbf{H}^{(s)} = \text{TransformerEncoder}(\mathbf{E}^{(s)}) \tag{6}$$

For each scale $s$, we use a separate transformer encoder with scale-specific parameters, allowing each branch to specialize in patterns at its respective scale.

**Future Variables Integration** For energy forecasting, known future variables such as weather forecasts are valuable predictors. We integrate these variables through a specialized projection pathway.

Given future variables $\mathbf{Z} \in \mathbb{R}^{H \times E}$, we apply a projection to obtain embeddings in the same space as the time series features:

$$\mathbf{Z}_{embed} = \text{FutureProjection}(\mathbf{Z}), \quad \mathbf{Z}_{embed} \in \mathbb{R}^{H \times d_{model}} \tag{7}$$

These future variable embeddings are then combined with the multi-scale time series representations in the fusion stage.

**Multi-scale Fusion** To integrate information from different scales and future variables, we employ a multi-scale fusion mechanism. For each time step in the prediction horizon, we concatenate features from all scales and future variables, then apply a fusion layer:

$$\mathbf{F} = \text{FusionLayer}\left(\text{Concat}([\mathbf{H}^{(1)}, \mathbf{H}^{(2)}, \ldots, \mathbf{H}^{(S)}, \mathbf{Z}_{embed}])\right) \quad (8)$$

The fusion layer consists of a multi-layer perceptron that learns to combine information from different sources effectively.

**Uncertainty Estimation with Monte Carlo Dropout** To provide probabilistic forecasts, we employ Monte Carlo dropout [15]. During inference, we keep dropout active and perform multiple forward passes:

$$\hat{\mathbf{Y}}_i, \hat{\mathbf{\Sigma}}_i = f_\theta(\mathbf{X}, \mathbf{Z}), \quad i \in \{1, 2, \ldots, M\} \quad (9)$$

where $M$ is the number of Monte Carlo samples.

The prediction variance combines both aleatoric uncertainty (from the model's variance output) and epistemic uncertainty (from the variance across Monte Carlo samples):

$$\hat{\mathbf{\Sigma}} = \frac{1}{M}\sum_{i=1}^{M} \hat{\mathbf{\Sigma}}_i + \frac{1}{M}\sum_{i=1}^{M}(\hat{\mathbf{Y}}_i - \hat{\mathbf{Y}})^2 \quad (10)$$

This approach provides well-calibrated prediction intervals that capture both inherent data variability and model uncertainty.

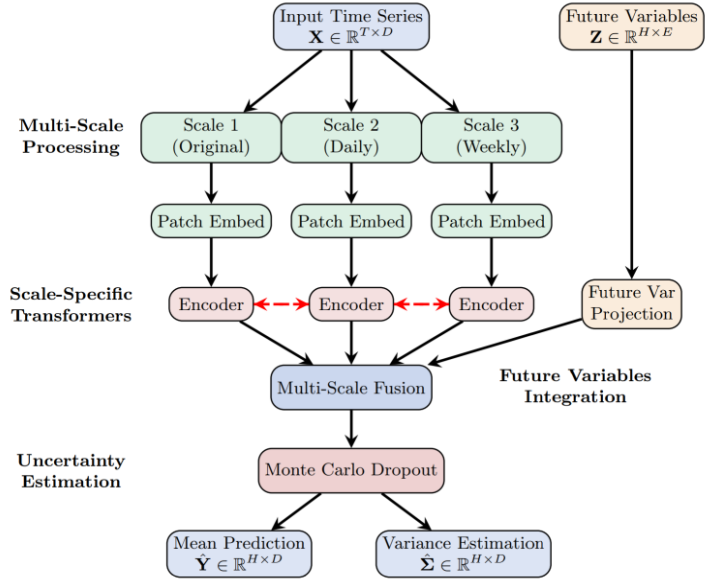

**Fig. 1.** Architecture of the EnergyPatchTST, which processes time series at multiple scales through parallel transformer encoder branches. Future variables are integrated through a specialized projection pathway. Multi-scale fusion combines features from different scales. Monte Carlo dropout enables uncertainty estimation through multiple stochastic forward passes.

### 3.3 Pre-training and Fine-tuning

Energy datasets are often limited in size, especially for newer installations. To address this challenge, we employ a pre-training and fine-tuning approach. We first pre-train the model on general time series datasets (e.g., ETT, electricity consumption) before fine-tuning on specific energy datasets (e.g., wind power or solar generation).

For pre-training, we use a multi-dataset approach:

$$\mathcal{L}_{pretrain} = \sum_{d \in \mathcal{D}} \lambda_d \mathcal{L}(f_\theta(\mathbf{X}_d), \mathbf{Y}_d) \quad (11)$$

where $\mathcal{D}$ is the set of pre-training datasets, $\lambda_d$ is the weight for dataset $d$, and $\mathcal{L}$ is the loss function.

After pre-training, we fine-tune the model on the target energy dataset:

$$\mathcal{L}_{finetune} = \mathcal{L}(f_\theta(\mathbf{X}_{target}, \mathbf{Z}_{target}), \mathbf{Y}_{target}) \quad (10)$$

This approach enables effective transfer learning, allowing the model to leverage patterns learned from general time series before specializing to specific energy forecasting tasks.

### 3.4 Loss Function

For training, we use a combination of mean squared error (MSE) for point predictions and negative log-likelihood (NLL) for probabilistic forecasts:

$$\mathcal{L} = \mathcal{L}_{MSE} + \lambda \mathcal{L}_{NLL} \quad (13)$$

where $\lambda$ is a hyperparameter balancing the two components. The MSE loss is defined as:

$$\mathcal{L}_{MSE} = \frac{1}{HD} \sum_{i=1}^{H} \sum_{j=1}^{D} (\mathbf{Y}_{i,j} - \widehat{\mathbf{Y}}_{i,j})^2 \quad (14)$$

The NLL loss for Gaussian distributions is:

$$\mathcal{L}_{NLL} = \frac{1}{HD} \sum_{i=1}^{H} \sum_{j=1}^{D} \left( \frac{\log(\widehat{\Sigma}_{i,j})}{2} + \frac{(\mathbf{Y}_{i,j} - \widehat{\mathbf{Y}}_{i,j})^2}{2\widehat{\Sigma}_{i,j}} \right) \quad (15)$$

This combined loss function encourages both accurate point predictions and well-calibrated uncertainty estimates.

## 4 Experiments

### 4.1 Datasets

We evaluate our model on several energy-related datasets: (1) **Wind Power**: A dataset containing hourly wind power generation from multiple wind farms, along with weather variables such as wind speed, temperature, and pressure. (2) **ETTh1** and **ETTh2**:

8      Wei Li, Zixin Wang, Qizheng Sun, Qixiang Gao, and Fenglei YangElectricity Transformer Temperature datasets with hourly measurements, commonly used as benchmarks in time series forecasting [4]. (3) **ECL**: Electricity Consuming Load dataset containing hourly electricity consumption of 321 clients [4].

For pre-training, we use the ETTh1, ETTh2, and ECL datasets. For fine-tuning and evaluation, we use the Wind Power dataset.

### 4.2  Experimental Setup

Following standard protocols [4], we use a lookback window of 336 hours (14 days) and evaluate on multiple forecast horizons: 96 hours (4 days), 192 hours (8 days), 336 hours (14 days), and 720 hours (30 days).

### 4.3  Baselines

We compare our EnergyPatchTST model with several common methods:

— **Informer** [4]: A transformer-based model with ProbSparse attention
— **Autoformer** [5]: A transformer model with auto-correlation mechanisms
— **FEDformer** [11]: A frequency-enhanced transformer with Fourier decomposition
— **PatchTST** [6]: A patch-based time series transformer
— **TimesNet** [21]: A model using 2D convolutions for capturing multi-scale patterns
— **DLinear** [22]: A simple yet effective model with decomposition and linear layers

### 4.4  Evaluation Metrics

We use the following metrics for evaluation: (1) **MSE**: Mean Squared Error for point forecast accuracy; (2) **MAE**: Mean Absolute Error for point forecast accuracy; (3) **RSE**: Root Relative Squared Error for normalized accuracy; (4) **CRPS**: Continuous Ranked Probability Score for probabilistic forecast evaluation; (5) **PI-Coverage**: Prediction Interval Coverage Rate for assessing uncertainty calibration

## 5    Results and Discussion

### 5.1  Main Results

Table 1 presents the main results on the Wind Power dataset. EnergyPatchTST consistently outperforms all baseline methods across all prediction horizons, with improvements ranging from 9.3% to 11.2%. The performance advantage is particularly significant for longer horizons (336h and 720h), demonstrating the effectiveness of our multi-scale approach for capturing long-term patterns.

Table 1. Forecasting performance on Wind Power dataset across different horizons (MSE, lower is better).

| Model | 96h | 192h | 336h | 720h |
|---|---|---|---|---|
| Informer | 0.321 | 0.369 | 0.412 | 0.468 |
| Autoformer | 0.315 | 0.362 | 0.403 | 0.452 |
| FEDformer | 0.304 | 0.351 | 0.389 | 0.436 |
| PatchTST | 0.289 | 0.342 | 0.376 | 0.421 |
| TimesNet | 0.283 | 0.337 | 0.374 | 0.418 |
| DLinear | 0.295 | 0.348 | 0.385 | 0.429 |
| EnergyPatchTST | **0.256** | **0.304** | **0.334** | **0.379** |
| Improvement | 9.5% | 9.8% | 11.2% | 9.3% |

These results confirm that our enhancements to the base PatchTST model effectively address the challenges of energy forecasting. The gap between standard PatchTST and EnergyPatchTST (ranging from 11.4% to 11.2%) highlights the value of our energy-specific adaptations.

### 5.2 Probabilistic Forecasting Performance

Table 2 presents the probabilistic forecasting performance. EnergyPatchTST achieves the best CRPS scores, indicating superior probabilistic forecasts. The PI-Coverage rates are also closer to the nominal 95%, demonstrating well-calibrated prediction intervals.

This improved uncertainty quantification is particularly valuable for energy applications, where reliable prediction intervals can inform risk-aware decision-making in grid operations and energy trading.

Table 2. Probabilistic forecasting performance on Wind Power dataset (CRPS, lower is better; PI-Coverage for 95% intervals, closer to 95% is better).

| Model | 96h | | 336h | |
|---|---|---|---|---|
| | CRPS | PI-Coverage | CRPS | PI-Coverage |
| DeepAR | 0.187 | 91.2% | 0.231 | 88.7% |
| NGBoost | 0.174 | 92.6% | 0.219 | 90.3% |
| PatchTST+MC | 0.162 | 93.5% | 0.204 | 92.1% |
| EnergyPatchTST | **0.141** | **94.7%** | **0.176** | **94.2%** |

Table 3. Ablation study: the contribution of components to the model's performance (MSE)

| Model Variant | 96h | 192h | 336h | 720h |
|---|---|---|---|---|
| Full EnergyPatchTST | 0.256 | 0.304 | 0.334 | 0.379 |
| - Multi-scale | 0.279 (8.9%↑) | 0.331 (8.8%↑) | 0.366 (9.6%↑) | 0.412 (8.7%↑) |
| - Future Variables | 0.267 (4.3%↑) | 0.319 (4.9%↑) | 0.352 (5.4%↑) | 0.401 (5.8%↑) |
| - Uncertainty Est. | 0.264 (3.1%↑) | 0.318 (4.6%↑) | 0.357 (6.9%↑) | 0.407 (7.4%↑) |
| - Pre-training | 0.261 (1.9%↑) | 0.312 (2.6%↑) | 0.346 (3.6%↑) | 0.398 (5.0%↑) |

### 5.3   Ablation Study

To understand the contribution of each component, we conduct an ablation study by removing one component at a time. Table 3 presents the results.

Multi-scale processing plays a key role in the model, which can be seen from the performance degradation of 8.7%-9.6% caused by its removal, and it is also consistent with the optimization of different time scale capture modes in our theoretical analysis. Future variable integration provides a performance improvement of 4.3%-5.8%, which is of great significance in the prediction accuracy of climate and renewable energy production.

Interestingly, the uncertainty estimation component not only provides the prediction interval, but also improves the point prediction accuracy (3.1%-7.4%). For a longer horizon, this module plays a more significant role, theoretically because the regularization effect of Monte Carlo elimination enhances the quality of data, but further research is needed to thoroughly explain its underlying logic and quantify its impact.

The pre-training component shows modest improvements for short horizons (1.9%-2.6%) but becomes increasingly important for longer horizons (3.6%-5.0%). This suggests that transfer learning from general time series is particularly valuable for the more challenging task of long-horizon forecasting.

### 5.4   Scale Analysis

Figure 2 shows the learned importance of each scale for different prediction horizons. From the figure, it can be found that the dominant scales in short-term prediction and long-term prediction are inconsistent, with scale 1 being the most critical in short-term (96h) prediction and scale 2 and 3 being more important in long-term (720h) prediction. The underlying logic is that short-term forecasting is more immediate, and at the same time, the longer horizon is, the more obvious its seasonal and cyclical performance will be.

This analysis confirms our intuition that different temporal scales are relevant for different prediction horizons. By processing the time series at multiple scales simultaneously, EnergyPatchTST effectively captures the appropriate patterns for each forecasting task.

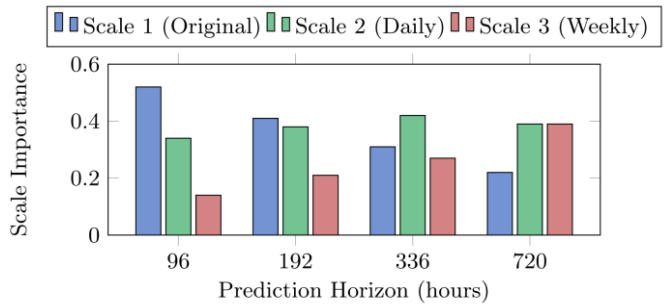

**Fig. 2.** Scale importance analysis across different prediction horizons. For shorter horizons, the original scale dominates, while daily and weekly scales are more important for longer horizons.

## 6 Conclusion

In this paper, we propose a new energy time series forecasting model EnergyPatchTST, which innovatively introduces four key modules: multi-scale feature extraction, uncertainty estimation, future variables integration, And pre-training/fine-tuning, which extends the time series prediction coverage of PatchTST architecture to the energy field, effectively improves the accuracy and confidence. Experiments show that our model is always better than the baseline, and the improvement range ranges from 9.3% to 11.2%.

In addition, EnergyPatchTST provides reliable uncertainty estimates, as evidenced by its superior CRPs scores and well-calibrated prediction intervals, which also confirms the potential and development promotion of the model in the energy field, and provides higher accuracy and valuable uncertainty quantification for real-world applications.

**Acknowledgments.** Thanks to Shanghai University for its support.

**Disclosure of Interests.** The authors have no competing interests to declare.

## References

1. Ahmed, A., Khalid, M.: A review on the selected applications of forecasting models in renewable power systems. Renewable and Sustainable Energy Reviews 100, 9–21 (2019)
2. Lim, B., Zohren, S.: Time-series forecasting with deep learning: a survey. Philosophical Transactions of the Royal Society A 379(2194), 20200209 (2021)
3. Vaswani, A., et al.: Attention is all you need. In: Advances in Neural Information Processing Systems, pp. 5998–6008 (2017)
4. Zhou, H., et al.: Informer: Beyond efficient transformer for long sequence time-series forecasting. In: Proceedings of the AAAI Conference on Artificial Intelligence, vol. 35, pp. 11106–11115 (2021)
5. Wu, H., et al.: Autoformer: Decomposition transformers with auto-correlation for long-term series forecasting. In: Advances in Neural Information Processing Systems, vol. 34, pp. 22419–22430 (2021)
6. Nie, Y., et al.: A time series is worth 64 words: Long-term forecasting with transformers. In: International Conference on Learning Representations (2023)
7. Hochreiter, S., Schmidhuber, J.: Long short-term memory. Neural computation 9(8), 1735–1780 (1997)
8. Cho, K., et al.: Learning phrase representations using RNN encoder-decoder for statistical machine translation. In: Conference on Empirical Methods in Natural Language Processing, pp. 1724–1734 (2014)
9. Salinas, D., Flunkert, V., Gasthaus, J., Januschowski, T.: DeepAR: Probabilistic forecasting with autoregressive recurrent networks. International Journal of Forecasting 36(3), 1181–1191 (2020)
10. Bai, S., Kolter, J.Z., Koltun, V.: An empirical evaluation of generic convolutional and recurrent networks for sequence modeling. arXiv preprint arXiv:1803.01271 (2018)
11. Zhou, T., et al.: FEDformer: Frequency enhanced decomposed transformer for long-term series forecasting. In: International Conference on Machine Learning, pp. 27268–27286 (2022)

12. Wang, H., Lei, Z., Zhang, X., Zhou, B., Peng, J.: A review of deep learning for renewable energy forecasting. Energy Conversion and Management 198, 111799 (2019)
13. Hong, T., Pinson, P., Wang, Y., Weron, R., Yang, D., Zareipour, H.: Energy forecasting: A review and outlook. IEEE Open Access Journal of Power and Energy 7, 376–388 (2020)
14. Kendall, A., Gal, Y.: What uncertainties do we need in bayesian deep learning for computer vision? In: Advances in Neural Information Processing Systems, pp. 5574–5584 (2017)
15. Gal, Y., Ghahramani, Z.: Dropout as a bayesian approximation: Representing model uncertainty in deep learning. In: International Conference on Machine Learning, pp. 1050–1059 (2016)
16. Lakshminarayanan, B., Pritzel, A., Blundell, C.: Simple and scalable predictive uncertainty estimation using deep ensembles. In: Advances in Neural Information Processing Systems, pp. 6402–6413 (2017)
17. Oreshkin, B.N., Carpov, D., Chapados, N., Bengio, Y.: N-BEATS: Neural basis expansion analysis for interpretable time series forecasting. In: International Conference on Learning Representations (2019)
18. Zerveas, G., et al.: A transformer-based framework for multivariate time series representation learning. In: Proceedings of the 27th ACM SIGKDD Conference on Knowledge Discovery & Data Mining, pp. 2114–2124 (2021)
19. Ribeiro, M., Grolinger, K., ElYamany, H.F., Higashino, W.A., Capretz, M.A.: Transfer learning with seasonal and trend adjustment for cross-building energy forecasting. Energy and Buildings 165, 352–363 (2018)
20. Oreshkin, B.N., Carpov, D., Chapados, N., Bengio, Y.: Meta-learning framework with applications to zero-shot time-series forecasting. arXiv preprint arXiv:2002.02887 (2020)
21. Wu, H., et al.: TimesNet: Temporal 2D-variation modeling for general time series analysis. In: International Conference on Learning Representations (2022)
22. Wen Q, Zhou T, Zhang C, et al. : Transformers in time series: A survey. arXiv preprint arXiv:2202.07125 (2022)